\documentclass[11pt,a4paper]{article}

\usepackage{graphicx}
\usepackage[export]{adjustbox}
\usepackage{arabtex}
\usepackage{utf8}
\usepackage[utf8x]{inputenc}

\usepackage{romannum}
\usepackage{array}
\usepackage{multicol}
\usepackage{comment}
\usepackage{caption}
\usepackage{subcaption}
\usepackage{multirow}

\usepackage[hyperref]{ranlp2021}

\hypersetup{pdfauthor   = Moustafa Al-Hajj ; Mustafa Jarrar,
            pdftitle    = ArabGlossBERT: Fine-Tuning BERT on Context-Gloss Pairs for WSD,
            pdfsubject  = RANLP 2021,
            pdfkeywords = {WSD ; Arabic ; lexicon ; Context-Gloss ;  NLP ; BERT ; Fine-tuning}
            }

\usepackage{times}
\usepackage{latexsym}

\usepackage{microtype}

\aclfinalcopy 

\def\correspondingauthor{\thanks{\ \ Corresponding author.}}

\title{ArabGlossBERT: Fine-Tuning BERT on Context-Gloss Pairs for WSD}

\author{Moustafa Al-Hajj \\
  Lebanese University\\
   Beirut, Lebanon \\
  \texttt{moustafa.alhajj@ul.edu.lb} \\
\And
  Mustafa Jarrar\correspondingauthor{} \\
  Birzeit University   \\
  Birzeit, Palestine \\
  \texttt{mjarrar@birzeit.edu} \\}

\date{}

\begin{document}
\setcode{utf8}
\maketitle
\begin{abstract}
Using pre-trained transformer models such as BERT has proven to be effective in many NLP tasks. This paper presents our work to fine-tune  BERT models for Arabic Word Sense Disambiguation (WSD). We treated the WSD task as a sentence-pair binary classification task. 
First, we constructed a dataset of labeled Arabic context-gloss pairs ($\sim$167k pairs) we extracted from the Arabic Ontology and the large lexicographic database available at Birzeit University. Each pair was labeled as $True$ or $False$ and target words in each context were identified and annotated. Second, we used this dataset for fine-tuning three pre-trained Arabic BERT models. Third, we experimented the use of different supervised signals used to emphasize target words in context. Our experiments achieved promising results (accuracy of $84\%$) although we used a large set of senses in the experiment.
\end{abstract}

\section{Introduction}
\label{sec:Introduction}
Word Sense Disambiguation (WSD) aims to determine which sense ({\em i.e.} meaning) a word may denote in a given context. This is a challenging task due to the semantic ambiguity of words. For example, the word ``book'' as a noun has ten different senses in Princeton WordNet such as ``a written work or composition that has been published'' and ``a number of pages bound together''. WSD has been a challenging task for many years but has gained recent attention due to the advances in contextualized word embedding models such as BERT \citep{devlin-etal-2019-bert}, ELMo \citep{peters-etal-2018-deep} and GPT-2 \citep{radford2018improving}. Such language models require less labeled training data since they are initially pre-trained on large corpora using self-supervised learning. The pre-trained language models can then be fine-tuned on various downstream NLP tasks such as sentiment analysis, social media mining, Named-Entity Recognition, word sense disambiguation, topic classification and summarization, among others.

A gloss is a short dictionary definition describing one sense of a lemma or lexical entry \cite{J06,J05}. A context is an example sentence in which the lemma or one of its inflections (\emph{i.e.} the target word) appears. In this paper, we aim to fine-tune Arabic models for Arabic WSD. Given a target word in a context and a set of glosses, we will fine-tune BERT models to decide which gloss is the correct sense of the target word. To do that, we converted the WSD task into a BERT sentence-pair binary classification task similar to \citep{huang2019,yap2020, blevins2020}. Thus, BERT is fine-tuned on a set of context-gloss pairs, where each pair is labeled as $True$ or $False$ to specify whether or not the gloss is the sense of the target word. In this way, the WSD task is converted into a sentence-pair classification task. 

One of the main challenges for fine-tuning BERT for Arabic WSD is that Arabic is a low-resourced language and that there are no proper labeled context-gloss datasets available. 

To overcome this challenge, we collected a relatively large set of definitions from the Arabic Ontology \cite{J21} and multiple Arabic dictionaries available at Birzeit University's lexicographic database \cite{JA19, JAM19} then we extracted glosses and contexts from lexicon definitions. 

Another challenge was to identify, locate and tag target words in context. Tagging target words with special markers is important in the fine-tuning phase because they act as supervised signals to highlight these words in their contexts, as will be explained in section \ref{sec:Methodology}. Identifying target words is not straightforward as they are typically inflections of lemmas, \emph{i.e.} with different spellings. Moreover, locating them is another challenge as the same word may appear multiple times in the same context with different senses. For example, the word  ({\scriptsize \<ذَهَب>}) appears two times in this  context ({\scriptsize \<ذَهَب ليشتري ذَهَب>}) with two different meanings: \emph{went} and \emph{gold}. We used several heuristics and techniques (as described in subsection \ref{sec:annotating}) to identify and locate target words in context in order to tag them with special markers. 

As a result, the dataset we constructed consists of about 167K context-gloss pair instances, 60K labeled as $True$ and 107K labeled as $False$. The dataset covers about 26k unique lemmas (undiacritized), 32K glosses and 60k contexts. 

We used this dataset to fine-tune three pre-trained Arabic BERT models: AraBERT \citep{arabert}, QARiB \citep{qarib} and CAMeLBERT \citep{inoue-etal-2021-interplay}\footnote{We were not able to use the ABERT and MARBERT models \cite{abdul2020arbert} as they appear very recently.}.  Each of the three models was fine-tuned for context-gloss binary classification. Furthermore, we investigated the use of different supervised signals used to highlight target words in context-gloss pairs.\\

The contributions of this paper can be summarized as follows:
\begin{enumerate}
\item Constructing a dataset of labeled Arabic context-gloss pairs; 
\item Identifying, locating and tagging target words;
\item Fine-tuning three BERT models for Arabic context-gloss pairs binary classification;
\item Investigating the use of different markers to highlight target words in context.  
 \end{enumerate}

The remainder of this paper is organized as follows. Section \ref{sec:relatedwork} presents related work. Section~\ref{sec:dataset} describes the constructed dataset and our methodology to extract and label context-gloss pairs, and splitting the dataset into training and testing sets. Section~\ref{sec:taskoverview} outlines the task we resolved in this paper and Section~\ref{sec:Methodology} presents the fine-tuning methodology. The experiments and the obtained results are presented in Sections~\ref{sec:experiment} and ~\ref{sec:results} respectively. Finally, Section~\ref{sec:conclusion} presents conclusions and future work.

\section{Related Work}
\label{sec:relatedwork}
Recent experiments in fine-tuning pre-trained language models for WSD and related tasks have shown promising results, especially those that use context-gloss pairs in fine-tuning such as \citep{huang2019, yap2020, blevins2020}. 

\citet{huang2019} proposed to fine-tune BERT on context-gloss pairs (\(label \in \{yes,no\}\)) for WSD, such that the gloss corresponding to the context-gloss pair candidate, with the highest output score for $yes$, is selected. \citet{yap2020} proposed to group context-gloss pairs with the same context but different candidate glosses as 1 training instance (groups of 4 and 6 instances). Then, they proposed to fine-tune BERT model on group instances with 1 neuron in the output layer. After that, they formulated WSD as a ranking/selection problem where the most probable sense is ranked first.

Others also suggested to emphasize target words in context-gloss training instances. \citet{huang2019,botha-etal-2020-entity,lei2017swim,yap2020} proposed to use different special signals in the training instance, which makes the target word ``special'' in it. As such, \citet{huang2019} proposed to use quotation marks around target words in context. In addition, they proposed to add the target word followed by a colon at the beginning of each gloss, which contributes to emphasizing the target word in the training instance. \citet{yap2020} proposed to surround the target word in context with two special [TGT] tokens. In contrast, \citet{botha-etal-2020-entity,lei2017swim} proposed to surround the target word in context with two different special tokens as marks of opening and closing. In this paper, we investigate the use of different types of signals to emphasize target words in context for Arabic WSD.

\citet{el2021arabic} fine-tuned two BERT models on a small dataset of context-gloss pairs, consisting of about 5k lemmas, about 15k positive and 15k negative context-gloss pairs. They claimed an F1-score of 89\%. However, this result is not reliable. After repeating the same experiment, we found that the majority of the context sentences used in the tests were already used for training. In this paper, we carefully selected the test set such that no contexts are used in both the training and the test sets. Additionally, we used a much larger sense repository (26k lemmas, 33k concepts and 167k context-gloss positive and negative pairs), which makes the task more challenging.

Other works related to Arabic WSD includes the use of static embeddings such as context and sense vectors \cite{laatar2017word}, Stem2Vec and Sense2Vec\cite{alkhatlan2018word}, Lemma2Vec \cite{al-hajj-jarrar-2021-lu},  Word Sense Induction \cite{alian2020sense}, or using fastText \cite{logacheva2020word}. \citet{elayeb2019arabic} reviewed Arabic WSD approaches until 2018.

\section{Dataset Construction} 
\label{sec:dataset}
This section describes how we constructed a dataset of labeled Arabic context-gloss pairs (See examples of pairs in Figure \ref{fig:pairs_example}). We extracted the context-gloss pairs from the Arabic Ontology and multiple lexicons in the Birzeit University's lexicographic database. The extracted pairs are labeled as $True$, and based on these $True$ pairs, we generated the $False$ pairs. Additionally, we identified the target word in each context and tagged it with different types of markers. 

\subsection{Context-Gloss Pairs Extraction}
\label{sec:trainigtestset}
Arabic is a low-resource language \cite{DH21} and there are no proper sense repositories available for Arabic \cite{KAJ21, JKKS21} that can be used to generate a dataset of context-gloss pairs, e.g. similar to the Princeton WordNet for English \cite{PWN}. The largest available lexical-semantic resource for Arabic is the Birzeit University's lexicographic database\footnote{Lexicographic Search Engine: \url{https://ontology.birzeit.edu/about}}, which contains the Arabic Ontology \citep{J21,J11} and about 400K glosses extracted from about 150 lexicons \citep{JA19, JAM19, ADJ19}. The problem is that each of the 150 lexicons covers a partial set of glosses and lemmas. Thus, for a given lemma, collecting the glosses from all lexicons may result in a set of redundant senses. Another problem is that some lexicons provide multiple senses within the same definition with no clear structure or separation markers, which makes it difficult to extract senses. Furthermore, some lexicons do not provide contexts (\emph{i.e.} example sentences) or they mix them with the definitions.\\

To overcome the above challenges and build a context-gloss pairs dataset, we performed the following steps:

\textbf{First, selection of candidate definitions:} We quired the 400K lexicon definitions to select a set of good candidate definitions. A good definition represents either one sense or multiple senses that are easy to parse and split (\emph{i.e.} contains some markers) and has context examples. That is, definitions that are not easy to parse or do not provide contexts were excluded.

\textbf{Second, extraction of glosses and contexts:} Each of the collected candidate definitions in the first phase was parsed and split into gloss(es) and context(s). Some definitions did not need to be split and some were split into separate glosses (one for each sense) in case a definition contains multiple glosses (\emph{i.e.} senses). Contexts were also extracted from the candidate definitions, taking into account that a definition may include multiple contexts for one sense. A parser was developed for each lexicon as each lexicon has its structure and text markers\footnote{We used the same parsing framework developed by \cite{ADJ19_report} for lexicon digitization.}. Nevertheless, some lexicons were clean and well-structured (e.g. the Arabic Ontology) that did not need any parsing.

\textbf{Third, selection of glosses and contexts:} Given that the glosses and contexts were extracted in the second phase, we applied the following criteria to select the glosses and contexts that we need to build a dataset of context-gloss pairs:

\begin{itemize} 
\item Short glosses and contexts (\emph{i.e.} one-word long) were excluded as they do not add useful information in the fine-tuning phase.

\item For each lemma, if one of its glosses does not have a context example then all glosses for this lemma were not selected. That is, for a lemma and its glosses to be selected, each gloss must have at least one context example.

\item In case the same lemma appears in multiple lexicons, the one with more glosses was selected. For example, let \emph{m} be a lemma with two glosses in lexicon A and three glosses in lexicon B, then the lexicon B set of glosses for \emph{m} is favored. If the same lemma has an equal number of glosses in multiple lexicons, we manually favor the more renowned lexicon. The idea of favoring lemmas with more glosses is because it indicates a richer set of distinct senses, and in this way, we avoid redundant senses for the same lemma in the dataset.

\item Only glosses for single-word lemmas are selected. Although multi-word expression lemmas are important, in this phase, we only focus on single-word lemmas as BERT can process single-word tokens. We plan to consider multi-word lemmas in the future.
\end{itemize}

\begin{table}
\begin{tabular}{ p{6cm}@{}>{\arraybackslash}m{3.5em}@{} } \hline 
 & \textbf{count} \\ \hline 
Unique Lemmas (undiacritized) & 26169 \\ 
Avg glosses per Lemmas & 1.25  \\ 
Unique Glosses & 32839 \\ 
Unique Contexts & 60272 \\ 
Avg context per gloss & 1.83 \\ 
True context-gloss pairs & 60323 \\ 
False context-gloss pairs & 106884 \\ 
Total True and False pairs & 167207 \\ 
\end{tabular}
\caption{\label{table:dataset_stat} Statistics about our context-gloss pairs dataset}
\end{table}

As a result, we selected about 32k glosses and 60k contexts for about 26K single lemmas (undiacritized), resulting in about 60k context-gloss pairs that we labeled as $True$ pairs (see Table \ref{table:dataset_stat} for more statistics). It is important to note that our dataset cannot be considered an Arabic sense repository because a sense repository should contain all senses for a given lemma, but our dataset does not necessarily include all senses for every lemma.

\subsection{Labeling Context-Gloss Pairs}
\label{sec:labeling_pairs}
The 60k context-gloss pairs extracted in the previous phase were labeled as $True$. The $False$ context-gloss pairs were then generated based on the $True$ pairs, as follows: For each lemma with more than one gloss, we cross-related its glosses with its contexts. For example, let ($context1-gloss1$) and ($context2-gloss2$) be the two $True$ pairs for the same lemma, then ($context1-gloss2$) and ($context2-gloss1$) are generated and labeled as $False$ pairs. As a result, about 107K context-gloss $False$ pairs were generated in this way.

\begin{figure*}[h]
    \centering
    \includegraphics[width=0.8\textwidth]{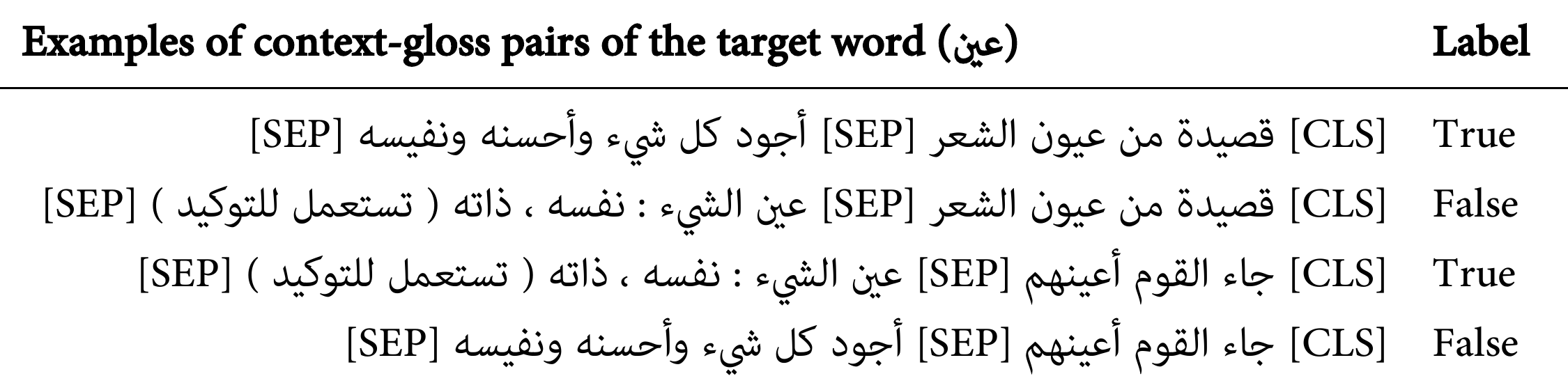}\par
    \caption{Examples of labeled context-gloss pairs}
    \label{fig:pairs_example}
\end{figure*}

\subsection{Annotating Target Words}
\label{sec:annotating}
This section presents our methodology for identifying the target word inside a given context and tagging it with a special supervised signal, which we need in the fine-tuning phase (see section \ref{sec:Methodology}). Figure \ref{fig:Signales} illustrates different tags of target words.

Given a lemma and a context, our goal is to identify which word is the target word in this context. As explained in section \ref{sec:Introduction}, a context is an example sentence in which a word (called target word) is mentioned with its sense defined in the gloss. Identifying a target word inside its context is not straightforward because: (i) it does not necessarily share the same spelling with its lemma, e.g. the word ({\scriptsize \<عيون>}) and its lemma ({\scriptsize \<عين>}) and, more importantly, (ii) it might occur multiple times and each time with a different sense such as ({\scriptsize \<كتب>}) which appears two times in this context ({\scriptsize \<كتب عدة كتب>}), with two different meanings: \emph{wrote} and \emph{books}.\\

The following four methods were performed at the same time to maximize the certainty in identifying target words. The resulting target words were verified manually:
\begin{itemize}

\item \emph{Sub-string}: We compared every word in the context with the given lemma (string-matching, after undiacritization). If the lemma is a sub-string of one or more words, then these words are candidate target words.

\item \emph{Character-level cosine similarity}: We developed a function\footnote{ The function converts two Arabic words (after removing diacritics) into two vectors (each cell represents the occurrence of a character), then computes their cosine similarity.} that takes a lemma and a context and returns the word with the max cosine similarity with the lemma. The minimum cosine value should be more than 0.75 $-$ an empirical threshold that we learned while reviewing the results. If a word is returned, then we considered it a candidate target word. 

\item \emph{Levenshtein distance}: This function takes a lemma and a context and returns the word with max Levenshtein distance (after removing diacritics) by comparing each word in the context with the lemma. The returned word is considered a candidate target word.

\item \emph{Lemmatization}: We used our in-house lemmatizer and lexicographic database to lemmatize every word in the given context and return those words that have their lemmas the same as the given lemma. The returned words are considered candidate target words.

\end{itemize}

These four methods were applied in parallel to maximize the certainty of correct matching and identification of target words. The results (candidate words, their scores and position) of the four methods were then combined and sorted (from more to less certain) and given to linguists to review. Each identified target word\footnote{In some cases, multiple words having the same sense can be considered target words inside the same context. For example ({\scriptsize \<كتابه>}) and ({\scriptsize \<الكتب>}) in the context ({\scriptsize \<كتابه كان من آفضل الكتب>}). In our dataset, we only considered one target word, most likely the first one.} was manually verified and, if needed, corrected by a linguist.

\subsection{Training and Test Datasets}
\label{sec:trainigtestset}
This section describes how we divided our dataset into training and test sets and the criteria we used to avoid repeated context in training and test sets. Recall that our dataset contains one or more glosses for each lemma and one or more contexts for each gloss, which we used to generate the context-gloss pairs dataset. The dataset cannot be arbitrarily divided as contexts used for training should not be used for testing. We selected the test set taking into account these two criteria: (\emph{i}) every context selected in the test set should not be selected in the training set and (\emph{ii}) every gloss should be selected in both the training and the test sets. 

Given these criteria, we selected the test set as follows:
(\emph{First}) we selected the pairs with repeated glosses from the set of context-gloss pairs (\emph{i.e.} glosses with more than one context). (\emph{Second}) we grouped pairs according to their glosses then selected one pair from each group larger than one and included it in the test set. All of these pairs were labeled as $True$. (\emph{Third}) we cross-related contexts with glosses of the same lemma to generate $False$ pairs in the test set from the $True$ pairs $-$ as described in subsection \ref{sec:labeling_pairs}. That is, again, the $False$ pairs were generated after selecting the $True$ pairs, and every pair selected for testing should not be part of the training set.

\begin{table}
\begin{tabular}{ |@{}>{\centering\arraybackslash}m{4em}@{}|lm{5.0em}@{}r|r|} \hline
 \textbf{Datasets} &  \textbf{Pairs} & \textbf{Count} & \textbf{Total} \\ \hline  \hline 
Training   & True pairs& 55,585 & \\ 
 & False pairs & 96,450  & 152,035 \\ \hline
Test  & True pairs & 4,738  & \\ 
 & False pairs & 10,434  & 15,172\\ \hline \hline 
 & & \textbf{Total}  & 167,207\\ \hline
\end{tabular}
\caption{\label{table:train_test_data} Counts of the training and testing pairs}
\end{table}

The resulted training and test datasets\footnote{The datasets and the fine-tuned BERT models are available at: \url{ https://ontology.birzeit.edu/downloads}} consist of 152,035 and 15,172 pairs, respectively. Table~\ref{table:train_test_data} provides statistics about the training and test sets.

\section{Task Overview}
\label{sec:taskoverview}
Given a context, a target word in the context and a gloss, our task is to decide whether or not the gloss corresponds to a specific sense of the target word. We approached the problem as a binary sequence-pair classification task. We concatenated the context and the gloss and separated them by the special [SEP] token (See Figure \ref{fig:pairs_example}). Afterward, we fine-tuned Arabic BERT models on our labeled dataset of context-gloss pairs (\(label \in \{True,False\}\)). \\

It is worth noting that although this binary context-gloss pair classification task is related to the WSD task, they are not exactly the same task. The WSD task aims at determining which sense (or gloss) a word in context denotes from a given set of senses. It is also worth noting that these two tasks are not the same as the Word-In-Context (WIC) task \cite{al-hajj-jarrar-2021-lu, martelli-etal-2021-semeval}, which aims at determining whether a target word has the same sense in two given contexts.

\section{Methodology}
\label{sec:Methodology}

\begin{figure*}[h]
    \centering
    \includegraphics[width=1\textwidth,frame]{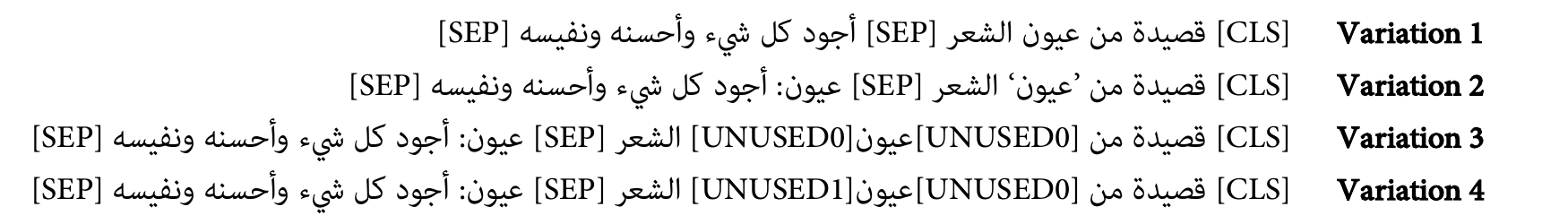}\par
    \caption{Illustration of the four context-gloss pairs variations.}
    \label{fig:Signales}
\end{figure*}

To address the binary context-gloss classification task, we experimented with four variations of the context-gloss pairs. The idea is to investigate using different supervised signals around target words to give them special attention during the fine-tuning. Figure~\ref{fig:Signales} illustrates these four variations. In variation 1, context-gloss pairs were left intact, without any signal. In the other three variations, we followed the techniques used by \citet{huang2019}, \citet{yap2020} and \citet{blevins2020} to signal target words. We surrounded target words with (\emph{\romannum{1}}) single quotes in variation 2, (\emph{\romannum{2}}) the special token [UNUSED0] in variation 3, and (\emph{\romannum{3}}) [UNUSED0] before and [UNUSED1] after in variation 4. Moreover, in the last three variations, we added the target word followed by a colon at the beginning of each gloss. In these four variations, the context and the gloss were concatenated into a sequence separated with the [SEP] token. \\

We fine-tuned three Arabic pre-trained models: AraBERT \cite{arabert}, QARiB \cite{qarib} and CAMeLBERT \cite{inoue-etal-2021-interplay} using our training dataset described in Section \ref{sec:dataset}. Before fine-tuning AraBERT, we used the pre-processing method used in \citep{arabert} to pre-train version 2 of their model. Before fine-tuning CAMeLBERT and QARiB models, we used the pre-processing method used in \citep{inoue-etal-2021-interplay} to pre-train the CAMeLBERT which consists in the normalization of alif maksura ({\scriptsize \<ى>}), teh marbuta ({\scriptsize \<ة>}), alif ({\scriptsize \<ا>}) and undiacritization.

Since BERT has a max length limit of tokens equal to 512, we limit the length of each training instance (\emph{i.e.} context-gloss pair) with a maximum of 512 tokens. Given, for example, the tokenizer used in AraBERTv02, only 216 pairs are larger than 512 tokens out of the 167,207 pairs in our dataset. Instances shorter than 512 were padded to the max length limit.

The BertForSequenceClassification model architecture is used in fine-tuning the three Arabic BERT models. The last hidden state of the token [CLS] is used for the classification task. The linear layer in the output consists of two neurons for the $True$ and $False$ classes. 

\section{Experiment Setup}
\label{sec:experiment}

We selected the base configuration of AraBERTv02, QARiB, and CAMeLBERT models due to computational constraints and as larger models do not necessitate better performance \citep{qarib,inoue-etal-2021-interplay}. We used the huggingface ``Trainer'' class in the fine-tuning. We performed a limited grid search to find a good hyperparameters combination then we fine-tuned each of the three models using the optimal configuration: initial learning rate of 2e-5, warmup\_steps of 1412 with a batch size of 16 over 4 training epochs. All other hyperparameters were kept at their default values. We used a single Tesla P100-PCIE-16GB in fine-tuning models.

\section{Results and Discussion}
\label{sec:results} 
This section presents the results of two experiments. Table \ref{table:results_modeles} presents the results of the first experiment in which we fine-tuned three BERT models on the variation 2 (\emph{i.e.} single quotes signal) of context-gloss pairs.

As AraBERTv02 outperformed other models in the first experiment, it has been chosen for conducting  a second experiment in which we fine-tuned on variation 1 (intact context-gloss pairs), variation 3 (two [UNUSED0] tokens around the target word in context-gloss pairs) and variation 4 ([UNUSED0] and [UNUSED1] tokens around the target word in context-gloss pairs). 
Reported results in Table \ref{table:results_signals} reveal that the use of different supervised signals around the target word did not significantly improve the overall results. The use of supervised signals reveals only $1\%$ of improvement over variation 1 (no signals). This improvement is comparable to the improvement of 1-2$\%$ achieved by \citet{huang2019} using special signals on English datasets.
 
\begin{table}[ht]
\centering
\begin{tabular}{|@{}>{\centering\arraybackslash}m{6.0em}@{}|l@{}>{\centering\arraybackslash}m{2.3em}@{}@{}>{\centering\arraybackslash}m{2.5em}@{}|@{}>{\centering\arraybackslash}m{4.2em}@{}|}
    \hline \textbf{Model}   &  &    \textbf{True} & \textbf{False} & \textbf{Accuracy} \\ \hline \hline
 
 \multirow{3}{*}{AraBERTv02}  & Precision & 81 & 85 & \multirow{3}{*}{84} \\ 
            & Recall & 66 & 93 &     \\ 
            & F1-score & 72 & 89 &     \\ \hline
 \multirow{3}{*}{CAMeLBERT}  & Precision & 77 & 83 & \multirow{3}{*}{82} \\  
            & Recall & 60 & 92 &     \\ 
            & F1-score & 67 & 87 &     \\ \hline
            
 \multirow{3}{*}{QARiB}  & Precision & 73 & 82 & \multirow{3}{*}{80} \\  
            & Recall & 58 & 90 &     \\ 
            & F1-score & 65 & 86 &     \\ \hline 
\end{tabular}
\caption{ Achieved results (\%) after fine-tuning three Arabic BERT models with the \emph{single quotes} supervised signal around the target word.}
\label{table:results_modeles}
\end{table}

\begin{table}[ht]
\centering
\begin{tabular}{|@{}>{\centering\arraybackslash}m{6.3em}@{}|l@{}>{\centering\arraybackslash}m{2.3em}@{}@{}>{\centering\arraybackslash}m{2.5em}@{}|@{}>{\centering\arraybackslash}m{4.2em}@{}|}
    \hline \textbf{Variation}   &  &    \textbf{True} & \textbf{False} & \textbf{Accuracy} \\ \hline \hline
 
  \multirow{3}{*}{\shortstack{\textbf{Variation 1} \vspace{0.3em} \\ No signal}}  & Precision & 80 & 85 & \multirow{3}{*}{83} \\  
           & Recall & 64 & 92 &     \\ 
            & F1-score & 71 & 88 &     \\ \hline
 
 \multirow{3}{*}{\shortstack{\textbf{Variation 3} \vspace{0.3em}\\ UNUSED0}}  & Precision & 81 & 85 & \multirow{3}{*}{84} \\  
            & Recall & 64 & 93 &     \\ 
            & F1-score & 71 & 89 &     \\ \hline

 \multirow{3}{*}{\shortstack{\textbf{Variation 4} \vspace{0.3em}\\UNUSED0,1}}  & Precision & 81 & 85 & \multirow{3}{*}{84} \\  
            & Recall & 64 & 93 &     \\
            & F1-score & 71 & 89 &     \\ \hline 
\end{tabular}
\caption{ Achieved results (\%) with AraBERTv02 using the other three supervised signals around the target word.}
\label{table:results_signals}
\end{table}

\section{Conclusion and Future Work}
\label{sec:conclusion}
We presented a large dataset of context-gloss pairs (167,207 pairs) that we carefully extracted from the Arabic Ontology and diverse lexicon definitions. Each pair was labeled as $True$ and $False$ and each target word in each context was annotated and tagged. We used this dataset to fine-tune three Arabic BERT models on binary context-gloss pair classification, and we achieved a promising accuracy of 84\%, especially as we used a large set of senses. Our experiments show that the use of different supervised signals around target words did not bring significant improvements (about $1\%$).\\

We will further build a large-scale content-gloss dataset. 
We also plan to include contexts written in Arabic dialects \cite{JHRAZ17} so that dialectal text can be sense-disambiguated. Additionally, we plan to consider Arabic text that is partially or fully diacritized, which requires lemmas across lexicons to be linked with each other \cite{JZAA18}. Lastly but more importantly, we plan to extend our work to address the WSD task and build a semantic analyzer for Arabic.

\section*{Acknowledgments}
We would like to thank the reviewers for their valuable comments and efforts for improving our manuscript. We would also like to thank Taymaa Hammouda for her technical support while preparing the dataset and annotating the contexts. We extend our thanks to Dr Abeer Naser Eddine for proofreading this paper.

\bibliographystyle{acl_natbib}
\bibliography{anthology,ranlp2021}
\end{document}